\pdfoutput=1

\documentclass[11pt]{article}

\usepackage{emnlp2021}

\usepackage{times}
\usepackage{graphicx}
\usepackage{subcaption}
\usepackage{amsmath}
\usepackage{latexsym}
\usepackage{multirow}
\usepackage{booktabs}

\usepackage[T1]{fontenc}

\usepackage[utf8]{inputenc}

\usepackage{microtype}

\usepackage{latexsym}

\usepackage{xspace}

\newcommand{\Sref}[1]{\S\ref{#1}}

\newcommand{\Fref}[1]{Figure~\ref{#1}}
\newcommand{\tref}[1]{table~\ref{#1}}

\newcommand{\sk}[1]{\textcolor{blue}{\textbf{[#1 -- SK]}}}

\newcommand{\ignore}[1]{}

\usepackage{microtype}

\newcommand{\ourmodel}{{\textsc{Para}vMF}\xspace}

\title{Improving the Diversity of Unsupervised Paraphrasing \\with Embedding Outputs}

\author{Monisha Jegadeesan\thanks{\ \ Currently at Google LLC} \\
  Indian Institute of Technology Madras \\
  \texttt{ monishaj.65@gmail} \\\And
  Sachin Kumar \\
   Carnegie Mellon University \\
  \texttt{sachink@cs.cmu.edu} \\\AND
  John Wieting\footnotemark[1] \\ 
   Carnegie Mellon University \\
  \texttt{jwieting@cs.cmu.edu} \\\And
  Yulia Tsvetkov \\
   University of Washington \\
  \texttt{yuliats@cs.washington.edu} \\}

\begin{document}
\maketitle
\begin{abstract}
We present a novel technique for zero-shot  paraphrase generation. 
The key contribution is an end-to-end multilingual paraphrasing model that is trained using translated parallel corpora to generate paraphrases into ``meaning spaces'' -- replacing the final softmax layer with word embeddings. 
This architectural modification, plus a training procedure that incorporates an autoencoding objective,
enables effective parameter sharing across languages for more fluent monolingual rewriting, and facilitates fluency and diversity in generation. 
Our continuous-output paraphrase generation models outperform zero-shot paraphrasing baselines, when evaluated on two languages using a battery of computational metrics as well as in human assessment.\footnote{The code is available at \url{https://github.com/monisha-jega/paraphrasing_embedding_outputs}} 
\end{abstract}

%
%
%
\section{Introduction}
\label{sect:introduction}
Paraphrasing aims to rewrite text while preserving its meaning and achieving a different surface realization. It is an eminently practical task, useful in educational applications \cite{inui2003text, petersen2007text, pavlick2016simple, xu2016optimizing}, information retrieval \cite{duboue2006answering, harabagiu2006methods, fader2014open}, 
in dialogue systems \cite{yan2016docchat}, 
as well as for data augmentation in a plethora of other tasks \cite{berant2014semantic, romano2006investigating, fadaee2017data, jin2018using, hou2018sequence}. 

Generating diverse and coherent paraphrases is a difficult task. Unlike in machine translation, where naturally occurring parallel data in the form of translated news, books and talks are available in abundance on the web, naturally occurring paraphrase corpora are scarce. Most common approaches to paraphrasing are based on translation, in the form of bilingual pivoting  \cite{bannard2005paraphrasing,mallinson2017paraphrasing} or back-translation \cite{wieting2017paranmt, hu2019parabank, hu2019large}. This stems from the hypothesis that if two sentences in a language (e.g.~English) have the same translation in another, (e.g.~French) they must be paraphrases of each other.
While these pipeline approaches bypass the problem of missing data, they propagate errors. 
Further, all neural paraphrasing models \cite[e.g.,][]{prakash2016neural, gupta2018deep, wang2019task} predict discrete tokens through a final softmax layer. We hypothesize that softmax-based architectures restrict the diversity of outputs, biasing the models to copy words and phrases from the input, which has an effect opposite to the intended one in paraphrasing. 

In this work, we introduce \ourmodel~-- a simple and effective method of training paraphrasing models by generating into embedding spaces (\Sref{sect:model}). Since parallel paraphrasing data is not available even in otherwise high-resource languages like French, we focus on an unsupervised approach. Using bilingual parallel corpora, we adapt multilingual machine translation \citep{johnson2017google} to monolingual translation. We propose to train this model with translation and autoencoding objectives. The latter helps simplify the training setup by using only one language pair, whereas prior work required multiple language pairs and more data to stabilize training \citep{tiedemann2018measuring,buck2018ask,guo2019zero, thompson2020paraphrase}. 
To encourage diversity, we propose to replace the final softmax layer in the decoder with a layer that learns to predict word vectors~\citep{kumar2018mises}. 
We show that predicting into word meaning representations increases diversity in paraphrasing by generating semantically similar words and phrases which are often neighbors in the embedding space. 

We evaluate our proposed model on paraphrasing English and French sentences (\Sref{sect:experiments}). In several setups, standard automatic metrics and human judgment experiments show that our zero-shot paraphrasing model with embedding outputs generates more diverse and fluent paraphrases, compared to state-of-the-art methods (\Sref{sect:results}). 

%

\section{The \ourmodel Model}
\label{sect:model}
Let the language to paraphrase in be $L_1$.
Our goal is to learn a mapping $f(\mathbf{x}; \theta)$ parameterized by $\theta$. $f$ takes a text $\mathbf{x} = (x_1, x_2, \cdots, x_m)$ containing $m$ words as input, which can be a sentence or a  segment in $L_1$. It then generates $\mathbf{y} = (y_1, y_2, \ldots, y_n)$ of length $n$ in the same language such that $\mathbf{x}$ and $\mathbf{y}$ are paraphrases. That is, $\mathbf{y}$ represents the same meaning as $\mathbf{x}$ using different phrasing. We assume that no direct supervision data is available, but there exists a bilingual parallel corpus between $L_1$ and another language $L_2$. We are also given pre-trained embeddings \cite{bojanowski2017enriching} for words in both $L_1$ and $L_2$. The dimension of both the embedding spaces is $d$. 

We use a standard transformer-based encoder-decoder model \citep{vaswani2017attention} as the underlying architecture for $f$. 
As visualized in 
the system diagram presented in the Appendix, $f$ is jointly trained to perform three tasks with a shared encoder and decoder: (1) translation from $L_1$ to $L_2$, (2) translation from $L_2$ to $L_1$ and (3) reconstructing the input text in $L_1$ (autoencoding).\footnote{To bias the model against always decoding in the other language, unlike in \citet{johnson2017google,tiedemann2018measuring}, we provide a language-specific start token in the encoder input, in addition to the decoder input.}

Towards our primary goal of meaning preservation, the translation objectives help the encoder map the inputs in both the languages to a common semantic space, whereas the decoder learns to generate language-specific outputs. On the other hand, with the autoencoding objective, we expose the model to examples where the input and output are in the same language, biasing the model to adhere to the start token supplied to it and decode monolingually.  
Using this training algorithm, we find in our experiments (\Sref{sect:results}), that the resulting paraphrases albeit meaning-preserving still lack in diversity. We identify two reasons for this issue. First, the model overfits to the autoencoding objective and just learns to copy the input sentences. We address this issue by using only a small random sample of the total training sentences for training with this objective.\footnote{We empirically determine this sample size to be ${\sim}1$\% of the total number of training examples.}

Second, we find that cross-entropy loss used to train the model results in peaky distributions at each decoding step where the target words get most of the probability mass. This distribution being another signal of overfitting also reduces diversity~\citep{meister-etal-2020-generalized}. We find in our preliminary experiments, that prior work to address this issue by augmenting diversity inducing objectives to the training loss~\citep{vijayakumar2018diverse} often comes at a cost of reducing meaning preservation. In this work, we propose using a different training loss which naturally promotes output diversity. We follow~\citet{kumar2018mises}, and instead of treating each word $w$ in the vocabulary as a discrete unit, we represent it using a unit-normalized pre-trained vector $\mathbf{e}$ learned using monolingual corpora~\citep{bojanowski2017enriching}.
At each decoding step, instead of predicting a probability distribution over the vocabulary using a softmax layer, we predict a $d$-dimensional continuous-valued vector $\mathbf{\hat{e}}$.  
We train our proposed model by minimizing von Mises-Fisher (vMF) loss---a probabilitistic variant of cosine distance---between the predicted vector and the pre-trained vector. 
At each step of decoding, the output word is generated by finding the closest neighbor (using cosine similarity) of the predicted output vector $\mathbf{\hat{e}}$ in the pre-trained embedding table. 
Since this loss does not directly optimize for a specific token but for a vector subspace which contains many similar meaning words, we observe that it has a higher tendency to generate diverse outputs than softmax-based models, both at the lexical and syntactic level as we show in our experiments. 

Overall, the contribution of this work is twofold: (1) a translation and autoencoding based training objective to enable paraphrasing while preserving meaning without any parallel paraphrasing data, and (2) optimizing for vector subspaces instead of token probabilities to induce diversity of outputs. 

\section{Experiments}
\label{sect:experiments}
\noindent
\textbf{Datasets \ }We evaluate paraphrasing in two languages: English and French. IWSLT'16 En$\leftrightarrow$Fr corpus \citep{cettolo2016iwslt} with ${\sim}220$K sentence pairs is used for training with translation objective, and $4450$ sentences, randomly sampled ${\sim}1$\% of the training data in $L_1$ (either En or Fr), for autoencoding. We use the $L_1$ side of the IWSLT'16 dev set for early stopping with the autoencoding objective.
We use IWSLT'16 test set for automatic evaluation consisting of $2331$ samples in En and Fr each. For human evaluation we subsample 200 sentences from this set.
We tokenize and truecase all the data using Moses preprocessing scripts \cite{koehn2007moses}. 
We conduct additional experiments with a larger En--Fr corpus constructed using a $2$M sentence-pair subset of the combination of the WMT'10 Gigaword \citep{TIEDEMANN12.463} and the OpenSubtitles corpora \citep{lison2016opensubtitles2016}.

\noindent
\textbf{Implementation \ }
We modify the standard seq2seq transformer model in OpenNMT~\citep{opennmt} to generate word embeddings \citep{kumar2018mises}, and train it with the vMF loss with respect to target vectors. 
We initialize and fix the input embeddings of the encoder and decoder with 
off-the-shelf (sub-word based) $\mathrm{fasttext}$ embeddings \citep{bojanowski2017enriching} for both En and Fr 
and align the embeddings to encourage cross-lingual sharing  \cite{artetxe2018robust}.
With a vocabulary size of $50$K for each language, the combined vocabulary size of the encoder and the decoder is $100$K.
Both encoder and decoder consist of $6$ layers with $4$ attention heads. 
The model is optimized using Adam \cite{kingma2015adam}, with batch size $4$K, and $0.3$ dropout. The 
hidden dimension size is $1024$, the dimension of the embedding layers is $512$. We add a linear layer to transform 300-dimensional pre-trained embeddings to 512-dimensional input vectors to the model. 
After decoding, we postprocess the generated output to replace words from $L_2$ by a look-up in the dictionary induced from the aligned embedding spaces.

\noindent
\textbf{Baselines \ }
   Although unsupervised methods of paraphrasing with only monolingual data have been explored in recent works~\citep{gupta2018deep, yang2019end, roy-grangier-2019-unsupervised, patro2018learning, park2019paraphrase} they have not been shown to outperform translation based baselines~\citep{west2020reflective}. Hence we compare our proposed approach with translation-based baselines only. First, we compare with
bilingual pivoting baselines  \cite{bannard2005paraphrasing,mallinson2017paraphrasing} which pipeline two separate translation models, $L_1\rightarrow L_2$, and $L_2 \rightarrow L_1$. We use two bilingual pivoting baselines, one based on continuous-output model (\textbf{\textsc{BP-vMF}}; the output vectors of the first model are first converted to discrete tokens before being fed to the next) and another based on softmax-based model (\textbf{\textsc{BP-CE}}).

To evaluate the impact of embedding outputs, we also compare our proposed model \textbf{\textsc{ParavMF}} to softmax-based baseline \textbf{\textsc{ParaCE}}, leaving other model components unchanged. \textbf{\textsc{ParaCE}} is a modified bilingual version of the multilingual method proposed in~\citet{guo2019zero}, the current state-of-the-art in zero-shot paraphrasing. 

\noindent
\textbf{Evaluation setup} 
There are many ways to paraphrase a sentence, but no manually crafted multi-reference paraphrase datasets exist, that could be used as test sets (and there are no datasets in languages other than English).
We thus evaluate the generated paraphrases on semantic similarity and lexical diversity compared to the \emph{input text}. 
Following prior work, we use the $n$-gram based metric 
\textbf{METEOR} \citep{banerjee2005meteor}.
Despite accounting for synonyms, it is not well-suited to evaluate paraphrases, since it typically assigns lower scores to novel phrasings, due to incomplete synonym dictionaries. We thus also include \textbf{BERTScore} \cite{zhang2019bertscore}, computing cosine similarity between the contextual embeddings of two sentences.
Naturally, just copying the inputs can also lead to high scores in these metrics. 
To evaluate lexical diversity, we follow \citet{hu2019large} and include \textbf{IoU} -- Intersection over Union (also called Jaccard Index) and Word Error Rate (\textbf{WER}). To measure structural diversity we use (constituency) Parse Tree Edit distance (\textbf{PTED}).\footnote{Before computing the PTED, we prune the tree to a max height of $3$, and discard all the terminal nodes. We employ Stanford CoreNLP \citep{manning2014stanford} for parsing and APTED algorithm for edit distance \citep{pawlik2015efficient}.} 
Note that model outputs that do not preserve meaning in paraphrasing (and generate totally different sentences) will also obtain high diversity scores, but these are not indicative of quality paraphrasing but will falsely contribute to high diversity scores if averaged across the entire test set. We thus measure the diversity only on subsets of the test set for which the strongest baseline (\textsc{Para}CE) and our model generate meaning-preserving paraphrases measured using BERTScore thresholds. We report the diversity scores for three such thresholds: $0.95$, $0.9$, $0.85$, selected empirically such that the sample size is sufficiently large.

\begin{table}
    \centering
    \begin{tabular}{l|cc | cc}
\toprule
\multirow{2}{*}{\textbf{Model}}& \multicolumn{2}{c|}{\textsc{English}} & \multicolumn{2}{c}{\textsc{French}}\\
& \textbf{BS$\uparrow$} & \textbf{MET.$\uparrow$} & \textbf{BS$\uparrow$} & \textbf{MET.$\uparrow$} \\
\midrule \midrule
\textsc{BP-CE} & 75.0 & 75.0 & 69.4 & 67.5\\
\textsc{BP-vMF} & 72.1 & 72.2 & 65.5 & 64.2 \\ 
\textsc{Para}CE & 83.5 & 87.4 & 82.3 & 81.6 \\
\ourmodel& \textbf{88.6} & \textbf{91.6} & \textbf{87.2} & \textbf{86.4 }\\ 
\bottomrule
\end{tabular}
    \caption{Meaning-preservation in generated paraphrases. BS: BertScore, MET: METEOR}
    \label{subtable:semantic}
\end{table}

\begin{table*}
\small 
\centering
\begin{tabular}{c|l|cccc|cccc}\toprule
\textbf{BERTScore} & \multirow{2}{*}{\textbf{Model}} &  \# (out & \multicolumn{3}{c|}{\textsc{English}} & \# (out  &\multicolumn{3}{c}{\textsc{French}} \\
\textbf{threshold} & & of 2K) & \textbf{IoU$\downarrow$} &\textbf{WER$\uparrow$} &\textbf{PTED$\uparrow$} & of 2K) & \textbf{IoU$\downarrow$} &\textbf{WER$\uparrow$} &\textbf{PTED$\uparrow$}\\
\midrule \midrule
0.85 & \textsc{Para}CE & 710 & 94.3          & 4            & \textbf{0.5} & 710 & 94.3 & 3.9 & \textbf{0.55}\\
     & \ourmodel       &     & \textbf{92.4} & \textbf{4.1} & 0.42         & & \textbf{92.7} & \textbf{4.1} & 0.42\\
\midrule
0.9  & \textsc{Para}CE & 539 & 96.2          & 2.6          & \textbf{0.34}  & 580 & 96.1 & 2.6 & \textbf{0.34}\\
     & \ourmodel       &     & \textbf{94.5} & \textbf{2.9} & 0.29          & & \textbf{94.5} & \textbf{2.9} & 0.29 \\
\midrule
0.95 & \textsc{Para}CE & 300 & 98.8          & 0.8          & 0.15          & 380 & 98.7 & 0.8 & 0.15\\
     & \ourmodel       &     & \textbf{97.7} & \textbf{1.2} & \textbf{0.16} & & \textbf{97.7} & \textbf{1.2} & \textbf{0.16} \\
\bottomrule
\end{tabular}
\caption{Diversity of meaning-preserving paraphrases compared to the test set. \ourmodel outperforms a strong baseline \textsc{Para}CE for both English and French, across all metrics for thresholds 0.85 and 0.9, and in IoU and WER for threshold of 0.95. 
}
\label{subtable:diversity}
\end{table*}

\section{Results}
\label{sect:results}

\textbf{Automatic evaluation} 
We observe in \tref{subtable:semantic} that \ourmodel outperforms all baselines in meaning-preservation. Both pivoting based baselines perform poorly on average. This is a consequence of error propagation exacerbated in \textsc{BP-vMF}\footnote{This is expected as \textsc{vMF} has been shown to slightly underperform CE for translation in prior work~\citep{kumar2018mises}. Our training procedure with an autoencoding objective alleviates this issue in \ourmodel.}. As a result, a very small fraction of generated sentences show meaning preservation (as measured by achieving a BERTScore greater than $0.85$). Hence, we only compare the diversity in the two best meaning-preserving models, \textbf{\textsc{ParaCE}} and \textbf{\textsc{ParaVMF}}. As shown in \tref{subtable:diversity}, across all thresholds the latter model achieves higher lexical and syntactic diversity in the outputs. Ablation results in the Appendix show that both the autoencoding objective and the final embedding layer contribute to the improved quality of paraphrases. 
An additional benefit of our proposed model is that by replacing the softmax layer with word embeddings, \ourmodel is trained 3x faster than the \textsc{ParaCE} baseline.

We further conduct a \textbf{manual evaluation} which quantifies the rate at which annotators find paraphrases fluent, consistent with input meaning, and novel in phrasing. In an A/B testing setup, we compare our proposed approach with the strongest baseline \textsc{ParaCE}.\footnote{Each judge is presented with a set of questions, each consisting of an input sentence and paraphrases generated by the two models as options, and is asked to choose the sentence that is fluent, meaning-preserving and offers a novel phrasing of the input. They are asked to choose neither if both sentences are dis-fluent and/or not able to preserve content. The options are shuffled.} 
200 sentences sampled from the IWSLT English test were scored by two annotators independently, which yielded the inter-annotator agreement of $0.37$ (fair agreement). 
Out of the sentences on which both annotators agree (142 out of 200), we find that
\ourmodel model outperforms the \textsc{Para}CE model in $73\%$ of votes. We show more details and some examples of \ourmodel and \textsc{ParaCE} system outputs in the Appendix.

Finally, we also evaluate that our results hold on a \textbf{larger dataset in different domain}. We retrain \ourmodel and \textsc{ParaCE} on $2M$ En--Fr corpus described in \Sref{sect:experiments}. \footnote{We use $4$K English sentences subsampled (${\sim}0.1$\% of the training data) from the same corpus for autoencoding. To further discourage copying, we use denoised autoencoding~\citep{lample2017unsupervised}.}
The results of automatic evaluation are presented in the Appendix.
We conduct human evaluation on a sample of 200 sentences from this test set following the same A/B testing procedure as described above, with each sample rated by three annotators, resulting in a pairwise-average kappa agreement index of 0.21.\footnote{We discarded around $53$ samples with no clear majority among the annotator ratings and report the results on the remaining samples, further ignoring cases where the paraphrases from both the models were rated to be of similar quality.} 42.9\% \ourmodel outputs were selected as better paraphrases, compared to 24.5\% outputs from \textsc{ParaCE}, supporting our main results on the IWSLT dataset. 
\section{Related Work} 
\label{sect:related}
Bilingual pivoting is a common technique used with bilingual data \citep{barzilay2001extracting,ganitkevitch2013ppdb, pavlick2015ppdb,bannard2005paraphrasing}. \textsc{Para}NMT \citep{wieting2017paranmt} is a large psuedo-parallel paraphrase corpus constructed through back-translation \citep{wieting2017learning}. \citet{iyyer2018adversarial} augment it with syntactic constraints for controlled paraphrasing; \textsc{Para}\textsc{Bank} \citep{hu2019parabank} improves upon \textsc{Para}NMT via lexical constraining of decoding; and \textsc{Para}\textsc{Bank}~2 \citep{hu2019large} improves the diversity of paraphrases in \textsc{Para}\textsc{Bank} through a clustering-based approach. Note that these works are focused on English. 
Here, we propose a language-independent approach relying only on abundant bilingual data.
Our approach is most similar to \citet{guo2019zero} who use bilingual and multilingual translation for zero-shot paraphrasing. They, however, observe that bilingual models are insufficient for paraphrasing and are often unable to produce the output in the correct language. We incorporate an autoencoding objective which simplifies and stabilizes training, and embedding-based outputs improving the diversity in paraphrasing.

\noindent
\section{Conclusion} 
\label{sect:conclusion}
We present \ourmodel, an end-to-end model for generating paraphrases, trained solely with bilingual data, without any paraphrase supervision. We propose to generate paraphrases into meaning spaces as opposed to discrete tokens. This leads to significant improvements in quality and diversity of paraphrasing over strong baselines.

\section*{Acknowledgments}
This material is based upon work supported by the National Science Foundation (NSF) under Grants No.~IIS2040926 and IIS2007960. The views and opinions of authors expressed herein do not necessarily state or reflect those of the NSF.

\bibliography{anthology,custom}

\begin{thebibliography}{53}
\expandafter\ifx\csname natexlab\endcsname\relax\def\natexlab#1{#1}\fi

\bibitem[{Artetxe et~al.(2018)Artetxe, Labaka, and Agirre}]{artetxe2018robust}
Mikel Artetxe, Gorka Labaka, and Eneko Agirre. 2018.
\newblock \href {https://www.aclweb.org/anthology/P18-1073} {A robust
  self-learning method for fully unsupervised cross-lingual mappings of word
  embeddings}.
\newblock In \emph{Proceedings of the 56th Annual Meeting of the Association
  for Computational Linguistics (Volume 1: Long Papers)}, pages 789--798,
  Melbourne, Australia.

\bibitem[{Banerjee and Lavie(2005)}]{banerjee2005meteor}
Satanjeev Banerjee and Alon Lavie. 2005.
\newblock {METEOR}: An automatic metric for mt evaluation with improved
  correlation with human judgments.
\newblock In \emph{Proceedings of the acl workshop on intrinsic and extrinsic
  evaluation measures for machine translation and/or summarization}, pages
  65--72.

\bibitem[{Barzilay and McKeown(2001)}]{barzilay2001extracting}
Regina Barzilay and Kathleen~R. McKeown. 2001.
\newblock \href {https://www.aclweb.org/anthology/P01-1008} {Extracting
  paraphrases from a parallel corpus}.
\newblock In \emph{Proceedings of the 39th Annual Meeting of the Association
  for Computational Linguistics}, pages 50--57.

\bibitem[{Berant and Liang(2014)}]{berant2014semantic}
Jonathan Berant and Percy Liang. 2014.
\newblock \href {https://www.aclweb.org/anthology/P14-1133} {Semantic parsing
  via paraphrasing}.
\newblock In \emph{Proceedings of the 52nd Annual Meeting of the Association
  for Computational Linguistics (Volume 1: Long Papers)}, pages 1415--1425.

\bibitem[{Bojanowski et~al.(2017)Bojanowski, Grave, Joulin, and
  Mikolov}]{bojanowski2017enriching}
Piotr Bojanowski, Edouard Grave, Armand Joulin, and Tomas Mikolov. 2017.
\newblock \href {https://www.aclweb.org/anthology/Q17-1010} {Enriching word
  vectors with subword information}.
\newblock \emph{Transactions of the Association for Computational Linguistics},
  5:135--146.

\bibitem[{Buck et~al.(2018)Buck, Bulian, Ciaramita, Gajewski, Gesmundo,
  Houlsby, and Wang.}]{buck2018ask}
Christian Buck, Jannis Bulian, Massimiliano Ciaramita, Wojciech Gajewski,
  Andrea Gesmundo, Neil Houlsby, and Wei Wang. 2018.
\newblock \href {https://openreview.net/forum?id=S1CChZ-CZ} {Ask the right
  questions: Active question reformulation with reinforcement learning}.
\newblock In \emph{International Conference on Learning Representations}.

\bibitem[{Cettolo et~al.(2016)Cettolo, Jan, Sebastian, Bentivogli, Cattoni, and
  Federico}]{cettolo2016iwslt}
Mauro Cettolo, Niehues Jan, St{\"u}ker Sebastian, Luisa Bentivogli, Roldano
  Cattoni, and Marcello Federico. 2016.
\newblock The iwslt 2016 evaluation campaign.
\newblock In \emph{International Workshop on Spoken Language Translation}.

\bibitem[{Duboue and Chu-Carroll(2006)}]{duboue2006answering}
Pablo Duboue and Jennifer Chu-Carroll. 2006.
\newblock \href {https://www.aclweb.org/anthology/N06-2009} {Answering the
  question you wish they had asked: The impact of paraphrasing for question
  answering}.
\newblock In \emph{Proceedings of the Human Language Technology Conference of
  the {NAACL}, Companion Volume: Short Papers}, pages 33--36. Association for
  Computational Linguistics.

\bibitem[{Fadaee et~al.(2017)Fadaee, Bisazza, and Monz}]{fadaee2017data}
Marzieh Fadaee, Arianna Bisazza, and Christof Monz. 2017.
\newblock \href {https://www.aclweb.org/anthology/P17-2090} {Data augmentation
  for low-resource neural machine translation}.
\newblock In \emph{Proceedings of the 55th Annual Meeting of the Association
  for Computational Linguistics (Volume 2: Short Papers)}, pages 567--573.

\bibitem[{Fader et~al.(2014)Fader, Zettlemoyer, and Etzioni}]{fader2014open}
Anthony Fader, Luke Zettlemoyer, and Oren Etzioni. 2014.
\newblock \href {https://doi.org/10.1145/2623330.2623677} {Open question
  answering over curated and extracted knowledge bases}.
\newblock In \emph{Proceedings of the 20th ACM SIGKDD International Conference
  on Knowledge Discovery and Data Mining}, page 1156–1165. Association for
  Computing Machinery.

\bibitem[{Ganitkevitch et~al.(2013)Ganitkevitch, Van~Durme, and
  Callison-Burch}]{ganitkevitch2013ppdb}
Juri Ganitkevitch, Benjamin Van~Durme, and Chris Callison-Burch. 2013.
\newblock \href {https://www.aclweb.org/anthology/N13-1092} {{PPDB}: The
  paraphrase database}.
\newblock In \emph{Proceedings of the 2013 Conference of the North {A}merican
  Chapter of the Association for Computational Linguistics: Human Language
  Technologies}, pages 758--764. Association for Computational Linguistics.

\bibitem[{Guo et~al.(2019)Guo, Liao, Jiang, Zhang, Zhang, and
  Liu}]{guo2019zero}
Yinpeng Guo, Yi~Liao, Xin Jiang, Qing Zhang, Yibo Zhang, and Qun Liu. 2019.
\newblock Zero-shot paraphrase generation with multilingual language models.
\newblock \emph{arXiv preprint arXiv:1911.03597}.

\bibitem[{Gupta et~al.(2018)Gupta, Agarwal, Singh, and Rai}]{gupta2018deep}
Ankush Gupta, Arvind Agarwal, Prawaan Singh, and Piyush Rai. 2018.
\newblock A deep generative framework for paraphrase generation.
\newblock In \emph{The Thirty-Second AAAI Conference on Artificial
  Intelligence}.

\bibitem[{Harabagiu and Hickl(2006)}]{harabagiu2006methods}
Sanda Harabagiu and Andrew Hickl. 2006.
\newblock \href {https://www.aclweb.org/anthology/P06-1114} {Methods for using
  textual entailment in open-domain question answering}.
\newblock In \emph{Proceedings of the 21st International Conference on
  Computational Linguistics and 44th Annual Meeting of the Association for
  Computational Linguistics}, pages 905--912. Association for Computational
  Linguistics.

\bibitem[{Hou et~al.(2018)Hou, Liu, Che, and Liu}]{hou2018sequence}
Yutai Hou, Yijia Liu, Wanxiang Che, and Ting Liu. 2018.
\newblock \href {https://www.aclweb.org/anthology/C18-1105}
  {Sequence-to-sequence data augmentation for dialogue language understanding}.
\newblock In \emph{Proceedings of the 27th International Conference on
  Computational Linguistics}, pages 1234--1245. Association for Computational
  Linguistics.

\bibitem[{Hu et~al.(2019{\natexlab{a}})Hu, Rudinger, Post, and
  Van~Durme}]{hu2019parabank}
J~Edward Hu, Rachel Rudinger, Matt Post, and Benjamin Van~Durme.
  2019{\natexlab{a}}.
\newblock \textsc{Para}\textsc{Bank}: Monolingual bitext generation and
  sentential paraphrasing via lexically-constrained neural machine translation.
\newblock In \emph{Proceedings of the AAAI Conference on Artificial
  Intelligence}, volume~33, pages 6521--6528.

\bibitem[{Hu et~al.(2019{\natexlab{b}})Hu, Singh, Holzenberger, Post, and
  Van~Durme}]{hu2019large}
J.~Edward Hu, Abhinav Singh, Nils Holzenberger, Matt Post, and Benjamin
  Van~Durme. 2019{\natexlab{b}}.
\newblock \href {https://www.aclweb.org/anthology/K19-1005} {Large-scale,
  diverse, paraphrastic bitexts via sampling and clustering}.
\newblock In \emph{Proceedings of the 23rd Conference on Computational Natural
  Language Learning (CoNLL)}, pages 44--54. Association for Computational
  Linguistics.

\bibitem[{Inui et~al.(2003)Inui, Fujita, Takahashi, Iida, and
  Iwakura}]{inui2003text}
Kentaro Inui, Atsushi Fujita, Tetsuro Takahashi, Ryu Iida, and Tomoya Iwakura.
  2003.
\newblock \href {https://doi.org/10.3115/1118984.1118986} {Text simplification
  for reading assistance: A project note}.
\newblock In \emph{Proceedings of the Second International Workshop on
  Paraphrasing - Volume 16}, page 9–16. Association for Computational
  Linguistics.

\bibitem[{Iyyer et~al.(2018)Iyyer, Wieting, Gimpel, and
  Zettlemoyer}]{iyyer2018adversarial}
Mohit Iyyer, John Wieting, Kevin Gimpel, and Luke Zettlemoyer. 2018.
\newblock \href {https://www.aclweb.org/anthology/N18-1170} {Adversarial
  example generation with syntactically controlled paraphrase networks}.
\newblock In \emph{Proceedings of the 2018 Conference of the North {A}merican
  Chapter of the Association for Computational Linguistics: Human Language
  Technologies, Volume 1 (Long Papers)}, pages 1875--1885.

\bibitem[{Jin et~al.(2018)Jin, King, Hussein, White, and
  Danforth}]{jin2018using}
Lifeng Jin, David King, Amad Hussein, Michael White, and Douglas Danforth.
  2018.
\newblock \href {https://www.aclweb.org/anthology/W18-0502} {Using paraphrasing
  and memory-augmented models to combat data sparsity in question
  interpretation with a virtual patient dialogue system}.
\newblock In \emph{Proceedings of the Thirteenth Workshop on Innovative Use of
  {NLP} for Building Educational Applications}, pages 13--23. Association for
  Computational Linguistics.

\bibitem[{Johnson et~al.(2017)Johnson, Schuster, Le, Krikun, Wu, Chen, Thorat,
  Vi{\'e}gas, Wattenberg, Corrado, Hughes, and Dean}]{johnson2017google}
Melvin Johnson, Mike Schuster, Quoc~V. Le, Maxim Krikun, Yonghui Wu, Zhifeng
  Chen, Nikhil Thorat, Fernanda Vi{\'e}gas, Martin Wattenberg, Greg Corrado,
  Macduff Hughes, and Jeffrey Dean. 2017.
\newblock \href {https://www.aclweb.org/anthology/Q17-1024} {{G}oogle{'}s
  multilingual neural machine translation system: Enabling zero-shot
  translation}.
\newblock volume~5, pages 339--351.

\bibitem[{Kingma and Ba(2015)}]{kingma2015adam}
Diederik~P Kingma and Jimmy Ba. 2015.
\newblock Adam: A method for stochastic optimization.
\newblock In \emph{International Conference on Learning Representations}.

\bibitem[{Klein et~al.(2017)Klein, Kim, Deng, Senellart, and Rush}]{opennmt}
Guillaume Klein, Yoon Kim, Yuntian Deng, Jean Senellart, and Alexander Rush.
  2017.
\newblock \href {https://www.aclweb.org/anthology/P17-4012} {{O}pen{NMT}:
  Open-source toolkit for neural machine translation}.
\newblock In \emph{Proceedings of {ACL} 2017, System Demonstrations}, pages
  67--72. Association for Computational Linguistics.

\bibitem[{Koehn et~al.(2007)Koehn, Hoang, Birch, Callison-Burch, Federico,
  Bertoldi, Cowan, Shen, Moran, Zens, Dyer, Bojar, Constantin, and
  Herbst}]{koehn2007moses}
Philipp Koehn, Hieu Hoang, Alexandra Birch, Chris Callison-Burch, Marcello
  Federico, Nicola Bertoldi, Brooke Cowan, Wade Shen, Christine Moran, Richard
  Zens, Chris Dyer, Ond\v{r}ej Bojar, Alexandra Constantin, and Evan Herbst.
  2007.
\newblock Moses: Open source toolkit for statistical machine translation.
\newblock In \emph{Proceedings of the 45th Annual Meeting of the ACL on
  Interactive Poster and Demonstration Sessions}, page 177–180. Association
  for Computational Linguistics.

\bibitem[{Kumar and Tsvetkov(2019)}]{kumar2018mises}
Sachin Kumar and Yulia Tsvetkov. 2019.
\newblock \href {https://openreview.net/forum?id=rJlDnoA5Y7} {Von
  {M}ises-{F}isher loss for training sequence to sequence models with
  continuous outputs}.
\newblock In \emph{International Conference on Learning Representations}.

\bibitem[{Lample et~al.(2018)Lample, Conneau, Denoyer, and
  Ranzato}]{lample2017unsupervised}
Guillaume Lample, Alexis Conneau, Ludovic Denoyer, and Marc'Aurelio Ranzato.
  2018.
\newblock \href {https://openreview.net/forum?id=rkYTTf-AZ} {Unsupervised
  machine translation using monolingual corpora only}.
\newblock In \emph{International Conference on Learning Representations}.

\bibitem[{Lison and Tiedemann(2016)}]{lison2016opensubtitles2016}
Pierre Lison and J{\"o}rg Tiedemann. 2016.
\newblock Opensubtitles2016: Extracting large parallel corpora from movie and
  tv subtitles.

\bibitem[{Mallinson et~al.(2017{\natexlab{a}})Mallinson, Sennrich, and
  Lapata}]{bannard2005paraphrasing}
Jonathan Mallinson, Rico Sennrich, and Mirella Lapata. 2017{\natexlab{a}}.
\newblock \href {https://www.aclweb.org/anthology/E17-1083} {Paraphrasing
  revisited with neural machine translation}.
\newblock In \emph{Proceedings of the 15th Conference of the {E}uropean Chapter
  of the Association for Computational Linguistics: Volume 1, Long Papers},
  pages 881--893.

\bibitem[{Mallinson et~al.(2017{\natexlab{b}})Mallinson, Sennrich, and
  Lapata}]{mallinson2017paraphrasing}
Jonathan Mallinson, Rico Sennrich, and Mirella Lapata. 2017{\natexlab{b}}.
\newblock \href {https://www.aclweb.org/anthology/E17-1083} {Paraphrasing
  revisited with neural machine translation}.
\newblock In \emph{Proceedings of the 15th Conference of the {E}uropean Chapter
  of the Association for Computational Linguistics: Volume 1, Long Papers},
  pages 881--893. Association for Computational Linguistics.

\bibitem[{Manning et~al.(2014)Manning, Surdeanu, Bauer, Finkel, Bethard, and
  McClosky}]{manning2014stanford}
Christopher Manning, Mihai Surdeanu, John Bauer, Jenny Finkel, Steven Bethard,
  and David McClosky. 2014.
\newblock \href {https://www.aclweb.org/anthology/P14-5010} {The {S}tanford
  {C}ore{NLP} natural language processing toolkit}.
\newblock In \emph{Proceedings of 52nd Annual Meeting of the Association for
  Computational Linguistics: System Demonstrations}, pages 55--60.

\bibitem[{Meister et~al.(2020)Meister, Salesky, and
  Cotterell}]{meister-etal-2020-generalized}
Clara Meister, Elizabeth Salesky, and Ryan Cotterell. 2020.
\newblock \href {https://doi.org/10.18653/v1/2020.acl-main.615} {Generalized
  entropy regularization or: There{'}s nothing special about label smoothing}.
\newblock In \emph{Proceedings of the 58th Annual Meeting of the Association
  for Computational Linguistics}, pages 6870--6886, Online. Association for
  Computational Linguistics.

\bibitem[{Park et~al.(2019)Park, Hwang, Chen, Choo, Ha, Kim, and
  Yim}]{park2019paraphrase}
Sunghyun Park, Seung-won Hwang, Fuxiang Chen, Jaegul Choo, Jung-Woo Ha, Sunghun
  Kim, and Jinyeong Yim. 2019.
\newblock Paraphrase diversification using counterfactual debiasing.
\newblock In \emph{Proceedings of the AAAI Conference on Artificial
  Intelligence}, volume~33, pages 6883--6891.

\bibitem[{Patro et~al.(2018)Patro, Kurmi, Kumar, and
  Namboodiri}]{patro2018learning}
Badri~Narayana Patro, Vinod~Kumar Kurmi, Sandeep Kumar, and Vinay Namboodiri.
  2018.
\newblock \href {https://www.aclweb.org/anthology/C18-1230} {Learning semantic
  sentence embeddings using sequential pair-wise discriminator}.
\newblock pages 2715--2729.

\bibitem[{Pavlick and Callison-Burch(2016)}]{pavlick2016simple}
Ellie Pavlick and Chris Callison-Burch. 2016.
\newblock \href {https://www.aclweb.org/anthology/P16-2024} {Simple {PPDB}: A
  paraphrase database for simplification}.
\newblock In \emph{Proceedings of the 54th Annual Meeting of the Association
  for Computational Linguistics (Volume 2: Short Papers)}, pages 143--148.

\bibitem[{Pavlick et~al.(2015)Pavlick, Rastogi, Ganitkevitch, Van~Durme, and
  Callison-Burch}]{pavlick2015ppdb}
Ellie Pavlick, Pushpendre Rastogi, Juri Ganitkevitch, Benjamin Van~Durme, and
  Chris Callison-Burch. 2015.
\newblock \href {https://www.aclweb.org/anthology/P15-2070} {{PPDB} 2.0: Better
  paraphrase ranking, fine-grained entailment relations, word embeddings, and
  style classification}.
\newblock In \emph{Proceedings of the 53rd Annual Meeting of the Association
  for Computational Linguistics and the 7th International Joint Conference on
  Natural Language Processing (Volume 2: Short Papers)}, pages 425--430.
  Association for Computational Linguistics.

\bibitem[{Pawlik and Augsten(2015)}]{pawlik2015efficient}
Mateusz Pawlik and Nikolaus Augsten. 2015.
\newblock \href {https://doi.org/10.1145/2699485} {Efficient computation of the
  tree edit distance}.
\newblock \emph{ACM Trans. Database Syst.}, 40(1).

\bibitem[{Petersen and Ostendorf(2007)}]{petersen2007text}
Sarah~E Petersen and Mari Ostendorf. 2007.
\newblock Text simplification for language learners: A corpus analysis.
\newblock In \emph{Workshop on Speech and Language Technology in Education}.

\bibitem[{Prakash et~al.(2016)Prakash, Hasan, Lee, Datla, Qadir, Liu, and
  Farri}]{prakash2016neural}
Aaditya Prakash, Sadid~A. Hasan, Kathy Lee, Vivek~V. Datla, Ashequl Qadir, Joey
  Liu, and Oladimeji Farri. 2016.
\newblock \href {http://arxiv.org/abs/1610.03098} {Neural paraphrase generation
  with stacked residual {LSTM} networks}.
\newblock \emph{CoRR}, abs/1610.03098.

\bibitem[{Romano et~al.(2006)Romano, Kouylekov, Szpektor, Dagan, and
  Lavelli}]{romano2006investigating}
Lorenza Romano, Milen Kouylekov, Idan Szpektor, Ido Dagan, and Alberto Lavelli.
  2006.
\newblock \href {https://www.aclweb.org/anthology/E06-1052} {Investigating a
  generic paraphrase-based approach for relation extraction}.
\newblock In \emph{11th Conference of the {E}uropean Chapter of the Association
  for Computational Linguistics}. Association for Computational Linguistics.

\bibitem[{Roy and Grangier(2019)}]{roy-grangier-2019-unsupervised}
Aurko Roy and David Grangier. 2019.
\newblock Unsupervised paraphrasing without translation.
\newblock In \emph{Proceedings of the 57th Annual Meeting of the Association
  for Computational Linguistics}.

\bibitem[{Thompson and Post(2020)}]{thompson2020paraphrase}
Brian Thompson and Matt Post. 2020.
\newblock Paraphrase generation as zero-shot multilingual translation:
  Disentangling semantic similarity from lexical and syntactic diversity.
\newblock \emph{arXiv preprint arXiv:2008.04935}.

\bibitem[{Tiedemann(2012)}]{TIEDEMANN12.463}
J{\"o}rg Tiedemann. 2012.
\newblock Parallel data, tools and interfaces in opus.
\newblock In \emph{Proceedings of the Eight International Conference on
  Language Resources and Evaluation (LREC'12)}, Istanbul, Turkey. European
  Language Resources Association (ELRA).

\bibitem[{Tiedemann and Scherrer(2019)}]{tiedemann2018measuring}
Jörg Tiedemann and Yves Scherrer. 2019.
\newblock \href {https://www.aclweb.org/anthology/W19-2005} {Measuring semantic
  abstraction of multilingual {NMT} with paraphrase recognition and generation
  tasks}.
\newblock In \emph{Proceedings of the 3rd Workshop on Evaluating Vector Space
  Representations for {NLP}}, pages 35--42.

\bibitem[{Vaswani et~al.(2017)Vaswani, Shazeer, Parmar, Uszkoreit, Jones,
  Gomez, Kaiser, and Polosukhin}]{vaswani2017attention}
Ashish Vaswani, Noam Shazeer, Niki Parmar, Jakob Uszkoreit, Llion Jones,
  Aidan~N. Gomez, undefinedukasz Kaiser, and Illia Polosukhin. 2017.
\newblock Attention is all you need.
\newblock In \emph{Proceedings of the 31st International Conference on Neural
  Information Processing Systems}, page 6000–6010.

\bibitem[{Vijayakumar et~al.(2018)Vijayakumar, Cogswell, Selvaraju, Sun, Lee,
  Crandall, and Batra}]{vijayakumar2018diverse}
Ashwin~K Vijayakumar, Michael Cogswell, Ramprasath~R. Selvaraju, Qing Sun,
  Stefan Lee, David Crandall, and Dhruv Batra. 2018.
\newblock \href {http://arxiv.org/abs/1610.02424} {Diverse beam search:
  Decoding diverse solutions from neural sequence models}.

\bibitem[{Wang et~al.(2019)Wang, Gupta, Chang, and Baldridge}]{wang2019task}
Su~Wang, Rahul Gupta, Nancy Chang, and Jason Baldridge. 2019.
\newblock A task in a suit and a tie: Paraphrase generation with semantic
  augmentation.
\newblock In \emph{Proceedings of the AAAI Conference on Artificial
  Intelligence}, volume~33, pages 7176--7183.

\bibitem[{West et~al.(2020)West, Lu, Holtzman, Bhagavatula, Hwang, and
  Choi}]{west2020reflective}
Peter West, Ximing Lu, Ari Holtzman, Chandra Bhagavatula, Jena Hwang, and Yejin
  Choi. 2020.
\newblock Reflective decoding: Unsupervised paraphrasing and abductive
  reasoning.
\newblock \emph{arXiv preprint arXiv:2010.08566}.

\bibitem[{Wieting and Gimpel(2018)}]{wieting2017paranmt}
John Wieting and Kevin Gimpel. 2018.
\newblock \href {https://www.aclweb.org/anthology/P18-1042}
  {\textsc{Para}{NMT}-50{M}: Pushing the limits of paraphrastic sentence
  embeddings with millions of machine translations}.
\newblock pages 451--462.

\bibitem[{Wieting et~al.(2017)Wieting, Mallinson, and
  Gimpel}]{wieting2017learning}
John Wieting, Jonathan Mallinson, and Kevin Gimpel. 2017.
\newblock \href {https://www.aclweb.org/anthology/D17-1026} {Learning
  paraphrastic sentence embeddings from back-translated bitext}.
\newblock In \emph{Proceedings of the 2017 Conference on Empirical Methods in
  Natural Language Processing}, pages 274--285. Association for Computational
  Linguistics.

\bibitem[{Xu et~al.(2016)Xu, Napoles, Pavlick, Chen, and
  Callison-Burch}]{xu2016optimizing}
Wei Xu, Courtney Napoles, Ellie Pavlick, Quanze Chen, and Chris Callison-Burch.
  2016.
\newblock \href {https://www.aclweb.org/anthology/Q16-1029} {Optimizing
  statistical machine translation for text simplification}.
\newblock \emph{Transactions of the Association for Computational Linguistics},
  4:401--415.

\bibitem[{Yan et~al.(2016)Yan, Duan, Bao, Chen, Zhou, Li, and
  Zhou}]{yan2016docchat}
Zhao Yan, Nan Duan, Junwei Bao, Peng Chen, Ming Zhou, Zhoujun Li, and Jianshe
  Zhou. 2016.
\newblock \href {https://www.aclweb.org/anthology/P16-1049} {{D}oc{C}hat: An
  information retrieval approach for chatbot engines using unstructured
  documents}.
\newblock In \emph{Proceedings of the 54th Annual Meeting of the Association
  for Computational Linguistics (Volume 1: Long Papers)}, pages 516--525.

\bibitem[{Yang et~al.(2019)Yang, Huo, Shen, Cheng, Wang, Wang, and
  Carin}]{yang2019end}
Qian Yang, Zhouyuan Huo, Dinghan Shen, Yong Cheng, Wenlin Wang, Guoyin Wang,
  and Lawrence Carin. 2019.
\newblock \href {https://www.aclweb.org/anthology/D19-1309} {An end-to-end
  generative architecture for paraphrase generation}.
\newblock In \emph{Proceedings of the 2019 Conference on Empirical Methods in
  Natural Language Processing and the 9th International Joint Conference on
  Natural Language Processing (EMNLP-IJCNLP)}, pages 3132--3142. Association
  for Computational Linguistics.

\bibitem[{Zhang et~al.(2020)Zhang, Kishore, Wu, Weinberger, and
  Artzi}]{zhang2019bertscore}
Tianyi Zhang, Varsha Kishore, Felix Wu, Kilian~Q. Weinberger, and Yoav Artzi.
  2020.
\newblock \href {https://openreview.net/forum?id=SkeHuCVFDr} {{BERTS}core:
  Evaluating text generation with bert}.
\newblock In \emph{International Conference on Learning Representations}.

\end{thebibliography}
\bibliographystyle{acl_natbib}

\appendix

\section{System diagram}
\begin{figure*}[th]
    \centering
     \begin{subfigure}[t]{0.49\textwidth}
     \includegraphics[scale=0.25]{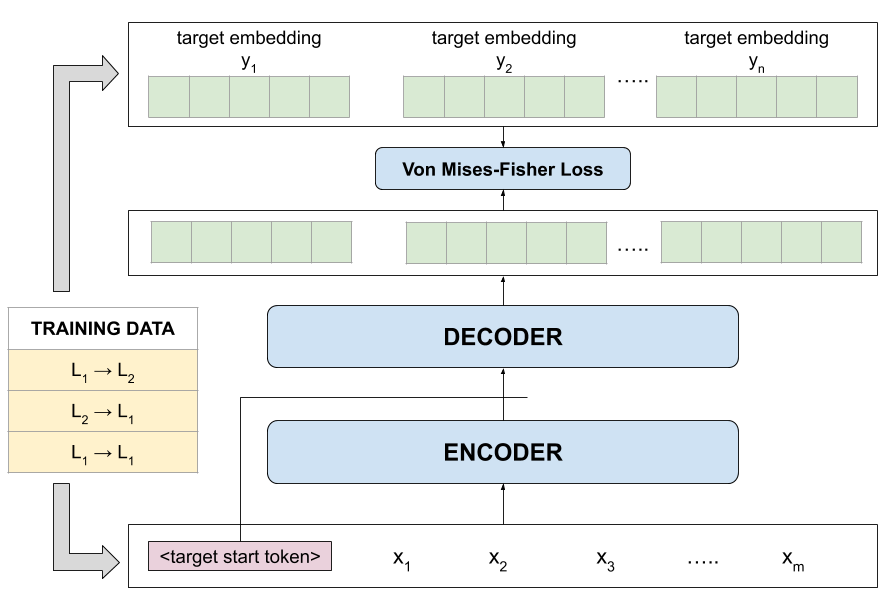}
     \caption{Training Procedure}
    \end{subfigure}
    \begin{subfigure}[t]{0.49\textwidth}
     \includegraphics[scale=0.25]{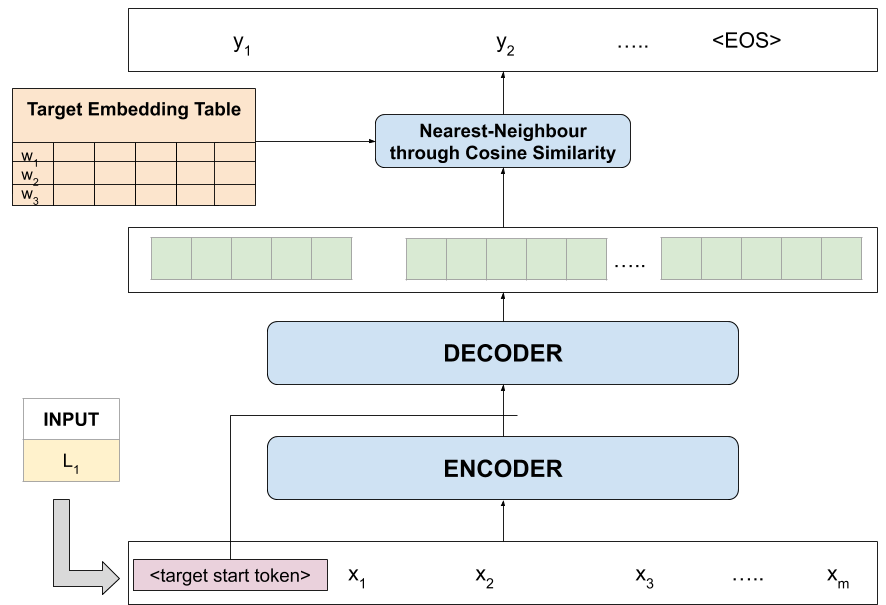}
     \caption{Testing Procedure}
    \end{subfigure}
    \caption{The \ourmodel Model: The decoder generates continuous-valued vectors at each step. It is trained by minimizing von Mises-Fisher loss between the output vectors and the pre-trained embeddings of the target words. Start tokens signalling the target language are supplied to both the encoder and the decoder. The training data consists of translation samples, $L_1\leftrightarrow L_2$ and autoencoding samples, $L_1 \rightarrow L_1$. During testing, the word in the target vocabulary whose embedding is closest to the generated output in terms of cosine similarity is output.
    }
    \label{fig:model}
\end{figure*}

The \ourmodel system is represented diagrammatically in \Fref{fig:model}.

\section{Example outputs}
 \label{app:outputs}
 Sample outputs of the \ourmodel and \textsc{Para}CE models are shown in \tref{table:outputs}.

\begin{table}
\begin{tabular}{c|ccc|cc}
\hline
& 
\multicolumn{2}{c}{Semantic Similarity}\\
\textbf{Model} 
& \textbf{BERTScore} & \textbf{METEOR}  \\
\hline \hline
\textsc{Para}NMT & 
61.6 & 62.1 \\ \hline
BP (vMF) & 
44.6 & 57.4 \\
BP (CE) & 
45.0 & 60.4\\
\textsc{Para}CE & 
65.9 & 81.7\\
\textbf{\ourmodel} &
\textbf{68.9} & \textbf{83.9}   \\
\hline
\end{tabular}
\caption{Evaluation of paraphrase generation on the \textsc{Para}NMT test set.}
\label{table:paranmteval}
\end{table}

\begin{table*}
\centering
\begin{subtable}{\textwidth}
\centering
\begin{tabular}{l|cc}
\toprule
\multirow{2}{*}{\textbf{Model}}& \multicolumn{2}{c}{\textsc{English}} \\
& \textbf{BERTScore$\uparrow$} & \textbf{METEOR$\uparrow$}\\
\midrule \midrule
\textsc{Para}CE & 62.2 & 73.6  \\
\ourmodel& \textbf{71.6} & \textbf{79.6}\\ 
\bottomrule
\end{tabular}
\caption{Semantic similarity between the test set and generated paraphrases}
\end{subtable}
\begin{subtable}{\textwidth}
\centering
\begin{tabular}{c|l|cccc}\toprule
\textbf{BERTScore} & \multirow{2}{*}{\textbf{Model}} & \# (out & \multicolumn{3}{c}{\textsc{English}} \\
\textbf{threshold} & & of 2K) & \textbf{IoU$\downarrow$} &\textbf{WER$\uparrow$} &\textbf{PTED$\uparrow$}\\
\midrule \midrule
0.85 & \textsc{Para}CE & 559 & 85.5          & 11.9            & \textbf{1.43}\\
     & \ourmodel       &     & \textbf{82.5} & \textbf{12.4} & 1.42    \\
\midrule
0.9  & \textsc{Para}CE & 327 & 91.2          & 7.0          & \textbf{0.81}\\
     & \ourmodel       &     & \textbf{87.9} & \textbf{8.3} & 0.64         \\
\midrule
0.95 & \textsc{Para}CE & 196 & 95.9          & 3.3          & 0.28      \\
     & \ourmodel       &     & \textbf{93.9} & \textbf{3.9} & \textbf{0.29} \\
\bottomrule
\end{tabular}
\caption{Diversity of meaning-preserving paraphrases compared to the test set}
\end{subtable}
\caption{Evaluation of paraphrase generation with \ourmodel  trained on $2M$ English-French sentence pairs. It outperforms a strong cross-entropy based baseline (\textsc{Para}CE) on semantic similarity and majority of diversity metrics. }
\label{table:bigeval}
\end{table*}

 \label{app:drawing}


\begin{table*}[!htp]\centering
\begin{tabular}{lll}\toprule
\toprule
Input &It 's expensive , it takes a long time , and it 's very complicated . \\
\toprule
\textsc{Para}CE &It 's expensive takes a time , and it 's very complicated . \\
\ourmodel &It 's costly , It takes a long time , and it 's very difficult . \\
\toprule
\toprule
Input &These are things to talk about and think about now , with your family and your loved ones . \\
\toprule
\textsc{Para}CE &These are things to talk about and think about now , with your family and your loved ones . \\
\ourmodel &These are things to speak of and think of now , with your family and the ones you love. \\
\toprule
\toprule
Input & So what opened my eyes ? \\
\toprule
\textsc{Para}CE & So what opened my eyes ? \\
\ourmodel & So what is it that opened my eyes up ? \\
\toprule
\toprule
Input & And this work has been wonderful . It 's been great . \\
\toprule
\textsc{Para}CE & And this work has been wonderful . It 's been great . \\
\ourmodel & This work has been wonderful and great . \\
\toprule
\toprule
Input & I wasn 't doing anything that was out of the ordinary at all . \\
\toprule
\textsc{Para}CE & I wasn 't doing anything that was out of the regular regular at all . \\
\ourmodel & I was doing nothing that was not ordinary . \\
\toprule
\toprule
Input & It will make tons of people watch , because people want this experience . \\
\toprule
\textsc{Para}CE & It will make tons of people watch , because people want this . \\
\ourmodel & Tonnes of people will look because they want this experience . \\
\bottomrule
\end{tabular}
\caption{Comparison of selected sample outputs for the IWSLT Test Set between \ourmodel model and the baselines. \ourmodel not only exhibits content preservation, but also demonstrates fluency as well as lexical and syntactic diversity.}
\label{table:outputs}
\end{table*}

\section{Training on a Larger Translation Dataset}
To measure the impact of the size of parallel translation data used for training, we conduct an experiment with a larger French-English corpus constructed using a $2M$ sentence-pair subset of the combination of the WMT'10 Gigaword \citep{TIEDEMANN12.463} and the OpenSubtitles corpora \citep{lison2016opensubtitles2016}. The semantic similarity scores and the diversity results are presented in \tref{table:bigeval}. The results of human evaluation are presented in the main paper.

\section{Evaluation on \textsc{Para}NMT-50M Test Set}
We evaluate the \ourmodel model (trained on English-French two-way translation data and English autoencoding data from the IWSLT'16 dataset) on test data sampled from \textsc{Para}NMT-50M \citep{wieting2017paranmt}, to demonstrate its paraphrasing ability on out-of-domain input, in addition to enabling direct comparison with back-translated data, as shown in \tref{table:paranmteval}. However, it is to be noted that the comparison is not a fair one, since \ourmodel is trained on just 220K data samples, wherease \textsc{Para}NMT is back-translated using a translation model that was trained on a bilingual dataset with a size of around $70M$.

\begin{table}[!htbp]
\centering
\begin{tabular}{c|c}
\midrule
\textbf{Model} & \textbf{Votes (\%)} \\
\midrule \midrule
\textsc{Para}CE & 39 (27.3\%)  \\
\ourmodel & \textbf{104} (\textbf{72.7\%}) \\
\midrule
\end{tabular}
\caption{\ourmodel outperforms the baseline in manual A/B testing (English).}
\label{table:humaneval}
\end{table}

\section{Ablation}
\label{sec:analysis}
\begin{table}
\centering
\begin{tabular}{@{}l|rrr}
\toprule
\textbf{Model}                                                 & \textbf{BLEU} & \textbf{BS} & \textbf{MET.} \\ \midrule
\ourmodel                                                        &  64.0               &       88.6             & 91.6                \\ 
- encoder start token     &  0.86         &      46.0              &         12.0                  \\
- autoencoding                                                  &              0.85     &          46.0          &  12.1          \\      \bottomrule
\end{tabular}
\caption{Performance of \ourmodel  without the proposed enhancements - removing either leads to a drastic performance drop}
\label{table:ablation}
\end{table}
We proposed three changes in a multilingual MT setup to use bilingual data for paraphrasing, (1) predicting continuous outputs and training with vMF loss, (2) language-specific start tokens in the encoder, and (3) an autoencoding objective. In the results section of the main paper, by comparing our method to \textsc{ParaCE}, we already established the importance of using vMF compared to cross-entropy. As shown in \tref{table:ablation}, ablating either of the other remaining two components leads to considerable performance drop. This is because the ablated models generate outputs in $L_2$ since they are never exposed to monolingual examples during training. Additional, in our preliminary experiments, we also observe that increasing the size of autoencoding data too much beyond ${\sim}1\%$ of the size of parallel translation data leads to a performance drop because the model just starts to learn to copy the input as-is rather than rephrasing.

\end{document}